\documentclass{article} 
\usepackage{iclr2025_conference,times}
\usepackage{tikz}
\usetikzlibrary{matrix,positioning}
\usepackage{booktabs}
\usepackage{multirow}
\usepackage{array}
\usepackage{tabularx}
\usepackage{makecell}

\usepackage{hyperref}
\usepackage{url}

\title{Measuring AI Agent Autonomy: Towards a Scalable Approach with Code Inspection}

\author{Peter Cihon \thanks{Equal contribution}  \\
GitHub\\
San Francisco, CA, USA \\
\texttt{petercihon@gmail.com} \\
\And
Merlin Stein $^{\ast}$
\\ University of Oxford \\
Oxford, UK \\
\texttt{merlin.stein@bsg.ox.ac.uk}
\And
Gagan Bansal \\
Microsoft \\
Redmond, WA, USA \\
\And
Sam Manning \\
Centre for the Governance of AI \\
Oxford, UK \\
\And
Kevin Xu \\
GitHub \\
San Francisco, CA, USA \\
}

\iclrfinalcopy 
\begin{document}

\maketitle
\begin{abstract}
AI agents are AI systems that can achieve complex goals autonomously. Assessing the level of agent autonomy is crucial for understanding both their potential benefits and risks. Current assessments of autonomy often focus on specific risks and rely on run-time evaluations -- observations of agent actions during operation. We introduce a code-based assessment of autonomy that eliminates the need to run an AI agent to perform specific tasks, thereby reducing the costs and risks associated with run-time evaluations. Using this code-based framework, the orchestration code used to run an AI agent can be scored according to a taxonomy that assesses attributes of autonomy: impact and oversight. We demonstrate this approach with the AutoGen framework and select applications.
\end{abstract}

\begin{figure}[h]
    \centering
        \caption{Towards a Complete Taxonomy of Autonomy for Agent Systems}
    \includegraphics[width=1\textwidth]{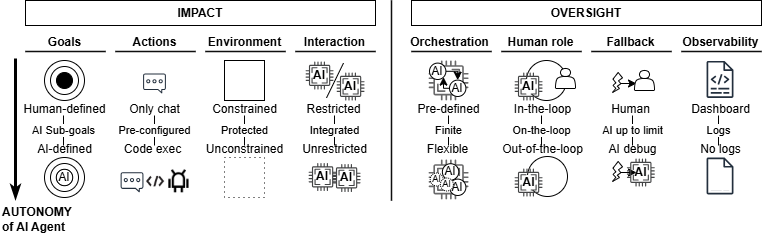}
    \label{fig:example}
\end{figure}

\section{Introduction}
Language model research and product attention focuses on creating Artificial Intelligence systems capable of flexibly planning and acting to influence environments over time (`AI agents') \citep{wang2024survey, kapoor2024}. In many cases, these systems orchestrate language models and their outputs to perform complex chains of thought and action for tasks ranging from software development to vacation planning. The responsible development and deployment of AI agents present many open questions today \citep{yo2023,chan2024,gabriel2024, wu2023autogen}. 

AI agents are designed to function autonomously. Autonomous systems can present risks of harm that have drawn policy scrutiny globally \citep{cihon2024,chan2023harms}. Early evaluations of AI agent capabilities have focused on general capability benchmarks \citep{agentbench,gaia,swebench}
and measuring specific threat models from frontier-capable models integrated into agent scaffolding \citep{autonomy_evals_guide, ukaisi}. 

Measuring the level of autonomy of an AI agent is helpful for a number of reasons. 

\begin{itemize}
    \item \textbf{Risk Assessment}: Measuring agent autonomy helps identify the range and consequence of decisions the agent might make, which is essential for assessing risks it might pose to those that interact with it.
    
    \item \textbf{Operational Control and Safe Deployment Strategies}: Understanding agent autonomy helps users and organizations set appropriate safety and control mechanisms, including escalation paths or fallback options. This understanding may influence deployment strategies such as phased roll outs, testing in restricted environments, or setting guardrails on certain AI agent behaviors prior to scaling up deployment.
    
    \item \textbf{Accountability, Liability, and Governance}: By measuring autonomy levels, organizations can more effectively allocate accountability and clarify liability across different actors, including users and AI agent developers.
    
\end{itemize}

This paper develops an initial method that could potentially be used to measure the autonomy of AI agents at scale. Complementing capability evaluations on running systems, our approach uses code inspection to assess how developers have structured the agent to function autonomously. Taking inspiration from 'autonomy levels' in other fields \citep{ISO_SAE_22736, yang2017medical}, our work aims to support research in developing empirically grounded measures of agent autonomy to inform policy and governance. 

Code-based evaluations offer several advantages over run-time evaluations but also come with certain limitations (see Table \ref{tab:code_runtime_comparison}). For policymakers, AI system developers, and evaluators, these evaluations can provide a resource-efficient, potentially scalable way to assess AI autonomy without the risks associated with run-time evaluations. This approach enables consistent, scaled evaluations that can help guide decisions, from establishing system security requirements to setting standards for safe AI deployment. Additionally, code inspection supports monitoring the development of autonomous AI agents in open source ecosystems and provides a framework for assessing proprietary applications when code access is available.

\begin{table}[h]
\caption{Code-Based Evaluations in Comparison to Run-Time Evaluations for Assessing Autonomy}
\label{tab:code_runtime_comparison}
\centering
\small
\begin{tabularx}{\textwidth}{@{}*{2}{>{\raggedright\arraybackslash}X}@{}}
\toprule
\multicolumn{1}{c}{\textbf{Benefits of Code-Based Evaluations}} & 
\multicolumn{1}{c}{\textbf{Limitations of Code-Based Evaluations}} \\
\midrule
\textbf{Reduced Risk of Harm During Evaluation:} 
Allows evaluation of autonomy features without running the AI agent, reducing potential for unintended, harmful behaviors during testing. & 
\textbf{Limited Insight into Emergent and Context-Specific Behaviors:} 
Cannot capture unexpected or adaptive behaviors that only emerge in real-world or complex environments, which may sometimes be observable with run-time evaluations. \\
\midrule
\textbf{Lower Evaluation Costs:} 
Avoids the need for resource-intensive run-time environments, leading to faster and more cost-effective autonomy assessments. &
\textbf{Challenges in Understanding User Interaction:} 
Misses real-time adjustments agents may make in response to user inputs, which can reveal nuances in autonomy that aren't evident from code alone. \\
\midrule
\textbf{Early Detection of Risk Factors:} 
Enables identification of autonomy risks early in development, allowing preemptive adjustments before deployment. & 
\textbf{Potential for Overlooking Latent Dependencies and Resource Usage:} 
Dependencies on external data sources or network availability can affect an AI agent autonomy under real operational conditions -- differences that may be more apparent at run-time. \\
\midrule
\textbf{Enables Scalable and Broad Evaluation:} 
Facilitates quick assessment across large numbers of agents or versions, enhancing comparability and consistency. & 
\textbf{Absence of Real-World Feedback Loops:} 
Code inspections cannot capture how agent actions may affect or alter the environment over time, creating feedback loops that might shift agent autonomy. \\
\bottomrule
\end{tabularx}
\end{table}

 Functioning agent systems make use of language models and `orchestration' or `scaffolding' software frameworks that routes model inputs and outputs \citep{davidson2023ai}. We focus on these software frameworks, seeking to answer (1) how frameworks, like AutoGen and LangChain, are designed for autonomy and (2) how do downstream developers implement these frameworks into applications for autonomy? As an initial proof of concept, we analyze AutoGen \citep{wu2023autogen} and some of its downstream applications. In subsequent work, this analysis could be scaled to possibly cover the wider universe of open source AI agent frameworks and applications. 
 
 Figure \ref{fig:example} summarizes our resulting conceptual framework of agent autonomy. This work makes three contributions: (1) a review of how system autonomy levels are operationalized in policy and AI (Section \ref{litreview}), (2) a taxonomy to assess autonomy levels of AI agents, based on AutoGen (Section \ref{Autogen}), and (3) a pilot code inspection rating the autonomy of select AI Agent applications (Section \ref{applications}). Section \ref{conclusion} concludes with implications, limitations and future work.  

\section{Approaches for assessing autonomy of systems} \label{litreview}
Autonomy has been defined and operationalized differently across contexts. Autonomy can include conceptions of self-actualization (constitutive) and capacity for interactions (behavioral) \citep{froese2007autonomy}. We focus on a form of the latter: \textit{decision autonomy}, which refers to the ability of an AI system to implement decisions without human oversight \citep{walsh2021autonomy} \footnote{This work is focused on informing governance of deployed AI systems. Thus, this paper builds on regulatory definitions of decision autonomy, leaving aside broader forms of autonomy or `agenticness' \citep{chan2023harms, ezenkwu2019machine}.} It is this concept that centers the impact of actions, environments and human oversight -- which are central for understanding and addressing risks from deployments.

Regulatory initiatives worldwide have started to define autonomy with a focus on \textit{decision autonomy}. The EU AI Act considers AI systems with varying degrees of autonomy, meaning that ``they have some degree of independence of actions from human involvement and of capabilities to operate without human intervention'' \citep{eu2024}. Similarly, \cite{NIST_SP1011} guidance defines a system as fully autonomous if it is ``expected to accomplish its mission, within a defined scope, without human intervention" and autonomy as levels characterized ``by factors including mission complexity and environmental difficulty.'' This guidance may be applicable across domains including manufacturing and national security \citep{nist_apply}.

In other industries, autonomy of systems is assessed along levels. For instance, autonomous driving levels are distinguished by the boundedness of the environment (\textit{operational design domain}), the responsibility for advanced actions (\textit{dynamic driving tasks}) and the degree of human oversight (\textit{fallback}) \citep{ISO_SAE_22736}. Similarly, degrees of autonomy of medical robots or aviation depend on the environment, difficulty of tasks and human oversight \citep{yang2017medical, anderson2018levels}.

For AI systems, researchers have operationalized levels of autonomy with empirical evaluations on tasks requiring increasing autonomy (benchmarked to humans taking minutes, hours, days or months to complete it) \citep{kinniment2023evaluating, morris2023levels}, according to levels of human oversight (in-, on-, and off-the-loop) \citep{simmler2021taxonomy}, autonomy of AI agents to determine suitable outputs and actions to take in the environment \citep{LangChainAgent2023, li2024personal}. Protocols have also been developed to evaluate an AI system's capability to pose autonomy-related risks \citep{autonomy_evals_guide}. Appendix \ref{autonomy-levels} provides an overview of levels of autonomy of different approaches.

In summary, across regulatory definitions, standards in other industries, and initial AI-specific characterisations, degrees of autonomy are differentiated based on (1) possible \textbf{impact} as a result of possible actions and environments, and (2) \textbf{oversight} in relation to orchestrating interactions and fallbacks within the system or between the system and the human. In the next section, we will operationalize an assessment of these attributes with AutoGen, an AI agent framework.

\section{Assessing autonomy of AutoGen}
\label{Autogen}

AutoGen is a popular\footnote{At the time of writing, the \href{https://github.com/microsoft/autogen}{AutoGen repository} had over 35000 stars, 5100 forks, and over 2300 dependents on GitHub. Please see this \href{https://x.com/pyautogen/status/1857264760951296210}{note} on governance changes in the open source project.} open source software framework for building AI agent systems using language models \citep{wu2023autogen}. AutoGen supports multi-agent conversations and tool-use to achieve arbitrary goals with varying levels of autonomy. It includes features that involve human users in the agent workflow, although downstream developers have ultimate control over how to configure their applications. Building on the literature above, we present a taxonomy of autonomy that assesses the specific features of the AutoGen framework for their relevance to system impact and oversight. For this preliminary work, we identify three general levels of autonomy associated with each attribute. For AutoGen, we focus on five most prominent attributes in Table \ref{autogen-table}. These attributes do not include others that are informed by the model layer and deployment monitoring, notably the granularity of goals set or pursued by the agent, inter-agent interaction, and fallbacks in deployment failure cases (See Figure \ref{fig:example}).

\begin{table}[h]
\caption{AutoGen-focused Taxonomy of Agent System Autonomy}
\label{autogen-table}
\centering
\small
\begin{tabularx}{\textwidth}{@{}l*{5}{>{\raggedright\arraybackslash}X}@{}}
\toprule
\multirow{3}{*}{\textbf{Autonomy}} & \multicolumn{2}{c}{\textbf{Impact}} & \multicolumn{3}{c}{\textbf{Oversight}} \\
\cmidrule(lr){2-3} \cmidrule(lr){4-6}
& \multicolumn{1}{c}{\makecell{Actions}} & 
\multicolumn{1}{c}{\makecell{Environment}} & 
\multicolumn{1}{c}{\makecell{Orchestration}} & \multicolumn{1}{c}{\makecell{Human-in-\\the-loop}} & \multicolumn{1}{c}{\makecell{Observability}} \\
\cmidrule(lr){2-3} \cmidrule(lr){4-6}
& \multicolumn{1}{c}{\footnotesize\itshape\makecell{What actions\\can the\\system take?}} & 
\multicolumn{1}{c}{\footnotesize\itshape\makecell{In what\\environment does\\the system operate?}} & 
\multicolumn{1}{c}{\footnotesize\itshape\makecell{How are agent\\interactions\\orchestrated?}} & 
\multicolumn{1}{c}{\footnotesize\itshape\makecell{How are humans\\kept in the loop?}} & 
\multicolumn{1}{c}{\footnotesize\itshape\makecell{How to monitor\\the system?}} \\
\midrule
Lower & None (conversation only) & Constrained (limits on internet access, memory, and compute) & Agent interaction is pre-determined & Agent(s) always consult human user & Dashboards or user-focused explanations \\
\\ & & & & & \\
Middle & Pre-configured tools for specific actions & Protected (Docker container deployment prevents agent from modifying its deployment environment directly)& Agent interaction is flexible but bounded (agent interactions are finite) & Agent(s) consult user for termination condition & Logs \\
\\ & & & & & \\
Higher & Code execution permits arbitrary actions & Unconstrained (access to both internet and local deployment) & Agent interaction is unbounded & Agent(s) never consult user & No logs\\
\bottomrule
\end{tabularx}
\end{table}

\textbf{IMPACT} concerns the overall consequences of what an agent system can do. The deployment of agent systems in critical or consequential use cases influences their impact. At the framework level (as opposed to specific application-level uses), impact is shaped by the system’s configuration, including the environment it can access and the range of actions it can take. 

\textbf{Actions} are the direct capabilities an agent has to influence an environment. AutoGen enables language models to take actions using tools, which may be more or less restricted. At its most constrained, AutoGen can facilitate agent conversation without any action. Other implementations can make use of registered tools, which can enable bounded actions (such as obtaining web content using a specified search engine, for example). When frameworks enable code execution, the potential for impactful actions increases significantly. This capability allows agents to execute any software-defined action, ranging from interacting with online resources to controlling physical actuators.

\textbf{Deployment environments} define the operational context and constraints for agent systems. By imposing access limits, they can reduce the potential impact of an agent’s actions. The AutoGen framework has two default configurations: local deployment or deployment to a Docker container. In the former, tools or executed code can alter the machine where the system is deployed, introducing computer security risks. The AutoGen framework default for code execution is the latter. By constraining the system’s environment to a Docker container, it cannot directly affect its deployment environment, even while it can access the broader internet (by default). Docker containers can be configured to limit network access, memory and compute resources. 

In sum, impact is a function of the interplay between the actions an agent can take and the (constrained) environment within which it operates. For example, a Docker-deployed system without access to tools represents a low-impact implementation, whereas a locally deployed system with unrestricted code execution exemplifies a high-impact configuration. 

\textbf{OVERSIGHT} concerns how developers and users may monitor and direct what an agent does. Design decisions at both the framework and application levels will impact how a user exercises oversight. At the framework level, these can be considered in three attributes: how agents are orchestrated, how the user is kept in the loop, and observability of the system during operation. 

\textbf{Orchestration} of agents within a system determines the flow of their interactions. The AutoGen framework supports single agent chains of tasks and multi-agent interactions. Both can be pre-determined in practice, where a developer designs a fixed flow for the agent system. Alternatively, agent interactions can be orchestrated flexibly, changing with each system run. Such flexibility is often bounded in practice: agent and multi-agent interaction (GroupChat) parameters govern an interaction, namely how it ends: for agents, how many auto-replies may be given and how many interactions total an agent may make, for GroupChat, how many rounds of interactions among all the agents. An interaction can be unbounded, where it does not have a set termination. At the extreme, AutoBuild\footnote{\url{https://microsoft.github.io/autogen/blog/2023/11/26/Agent-AutoBuild/}} is an implementation that can permit the unbounded creation of sub-agents to complete a task.

\textbf{Human-in-the-loop} is an express feature of the AutoGen framework. Each agent has a parameter `human\_input\_mode' which can take three values: never, terminate, and always. With `always,' an agent will seek input from a human following every step, though it may continue after a delay insofar as the agent was initialized with an `auto-reply'; else, the agent will be blocked awaiting human input. For `terminate', the agent will continue until given the terminate command. With `never', the agent will never wait for human input once initialized. `Never' may be common in systems with a Society of Mind agent or otherwise Nested Chat: for example, an user-agent may interact with the human while running numerous sub-agents without direct human input.\footnote{\url{https://microsoft.github.io/autogen/docs/reference/agentchat/contrib/society_of_mind_agent/}} In practice, this may complicate analyzing oversight by simply looking for the presence of a human in the loop, when a system could do so while retaining complex and possibly opaque autonomous agent interactions. Understanding the needs and competence of intended users can support a better assessment of humans-in-the-loop, although this is beyond the scope of our code inspection method.

\textbf{Observability} means that a user can see what an agent system is doing. The AutoGen framework supports logging, documenting each time an agent within the system is invoked, the input, output, time, cost, and associated identifiers. Logs may be of use to developers looking to debug an application they are building. However, if the logs are not used further to create explanations or dashboards, end-users may not have meaningful observability into agent actions.

See Table \ref{autogen-table} for a summary. These attributes are not exhaustive: other aspects of agent frameworks, downstream application development, and use influence autonomy of systems in practice. Additional considerations are raised at the model layer and/or deployment time, including the specificty of goals pursued by the agent, its interactions with other agents, and means of falling-back in failure cases. These relate to an agent taking initiative (while considering costs and benefits of interruptions, user effort, and the chance of getting meaningful user input \citep{horvitz1999principles}). Our approach is practical rather than comprehensive: to facilitate empirical evaluation of applications being built with agent frameworks. We offer a case study assessment for AutoGen to provide an illustrative example of how the autonomy taxonomy can be applied to agent frameworks more broadly.

\section{Assessing autonomy of AutoGen applications}
\label{applications}
 We operationalize the autonomy taxonomy by inspecting the source code for AutoGen applications. Appendix \ref{code} describes code flags used for scoring. Ten AutoGen applications\footnote{Top 5 dependent repositories identified from \href{https://github.com/microsoft/autogen}{AutoGen repository} by stars, and 5 uniquely significant repositories like AutoGen Studio.} were scored by three researchers. Consensus results are shown in Table \ref{app-table} and further detail in Appendix \ref{code}. 
\begin{table}
\caption{Scoring Autonomy of Selected AutoGen Applications}
\label{app-table}
\centering
\small
\begin{tabularx}{\textwidth}{@{}l*{5}{>{\raggedright\arraybackslash}X}@{}}
\toprule
\multirow{3}{*}{\textbf{Autonomy}} & \multicolumn{2}{c}{\textbf{Impact}} & \multicolumn{3}{c}{\textbf{Oversight}} \\
\cmidrule(lr){2-3} \cmidrule(lr){4-6}
& \multicolumn{1}{c}{\makecell{Actions}} & 
\multicolumn{1}{c}{\makecell{Environment}} & 
\multicolumn{1}{c}{\makecell{Orchestration}} & \multicolumn{1}{c}{\makecell{Human-in-\\the-loop}} & \multicolumn{1}{c}{\makecell{Observability}} \\
\midrule
Lower & -- & 5 & 2, 5 & 5, 10 & 2, 4, 7, 8, 9 \\
\\ & & & & & \\
Middle & 5, 6 & 4, 6 & 1, 3, 4, 6, 8 & 8 & 1, 3, 6, 5, 10 \\
\\ & & & & & \\
Higher & 1, 2, 3, 4, 7, 8, 9, 10 & 1, 2, 3, 7, 8, 9, 10 & 7, 9, 10 & 1, 2, 3, 4, 6, 7, 9 & -- \\
\bottomrule
\end{tabularx}

\vspace{0.5em}
\raggedright
\small
\textbf{Key:} 1: \href{https://github.com/microsoft/autogen/tree/5e0b677acc10bbbf4fab889bbcc0c70f3f13abb8/python/packages/autogen-studio}{AutoGen Studio}, 2: \href{https://github.com/ComposioHQ/composio}{Composio}, 3: \href{https://github.com/Ag2S1/Sibyl-System}{Sibyl System}, 4: \href{https://github.com/yanivvak/dream-team}{Dream Team}, 5: \href{https://github.com/karthik-codex/Autogen_GraphRAG_Ollama}{GraphRag\_Ollama}, 6: \href{https://github.com/agentcoinorg/AutoTx}{AutoTx}, 7: \href{https://github.com/letta-ai/letta}{Letta}, 8: \href{https://github.com/h2oai/h2ogpt}{h2oGPT}, 9: \href{https://github.com/langflow-ai/langflow}{Langflow}, 10: \href{https://github.com/binary-husky/gpt_academic}{GPT-Academic}
\end{table}

The analysis shows that a differentiated categorization and scoring of agent autonomy with code inspections is tractable. The three raters demonstrated substantial inter-rater agreement (Fleiss' $k = 0.64$) across all evaluation categories, indicating consistent application of the rating criteria; additional agreement measures reported in Appendix \ref{code}.

The variations between AutoGen applications are notable. Flexible platform applications like AutoGen Studio or Composio allow users to build specific applications with multiple degrees of autonomy. For example, this includes code executions, inside or outside pre-specified environments like Docker containers. Applications range from pre-configured machine learning agent with multiple sub-agents who build and critique code, to financial research agent and editing agent teams.

Specific applications like Dream Team for development with quality assurance or GraphRAG for data-based reasoning are more constrained for most autonomy attributes. Potentially for their reliability for a specific purpose, they operate in bounded Docker environments. However, both applications still spin up one subagent that requires a human-in-the-loop, as a fallback or main input, and the remaining -- executing -- subagents human-off-the-loop. 

This pilot has limitations, partly due to our small sample and partly due to the nature of code inspections. Inter-rater agreement varied by attribute: orchestration ($k = 0.67$), human-in-the-loop ($k = 0.65$), environment ($k = 0.60$), observability ($k = 0.47$), and action ($k = 0.30$), with the latter three being below the substantial agreement threshold. Actions may be enabled not only by the AutoGen framework but by downstream developer choices to make custom tools or use additional frameworks, complicating review. Observability proved challenging to consistently differentiate between maintaining logs that may be useful to developers and meaningful awareness for users that may limit autonomy in practice. Consistent assessment of the constrained environment also proved challenging, where in some cases wider access may be possible through API calls to other systems. Inter-rater agreements aide, additional limitations apply to the human-in-the-loop attribute. Is an agent that requires human approval only after trying multiple different information retrieval tools more or less autonomous than an agent that requires human approval after each mouse click? It might depend on more granular assessments of the kind of actions before which human approval is required.

No agent system received lower scores on actions and observability. The first may be explained by the common interest in developing agent systems to take actions, i.e., to do more than simple chat interactions that are also  possible with AutoGen. The open-source nature and focus on developers of the inspected agent systems, might lead to a focus on transparency and consequently logs. Closed-source systems might still offer logs to developers, but not to users.  

\section{Conclusion and future work}
\label{conclusion}
This paper piloted a code-inspection approach to assessing agent systems for their level of autonomy. With further development as identified below, this approach holds promise for agent assessments at scale. Code inspection rates AI agent applications without running them, which is resource-efficient, reduces risks associated with run-time evaluations, and can enable uniform, scaled assessments that can inform stakeholder decisions -- from model developers to policymakers. It also supports risk assessment to inform what level of system security might be important before running them. This method could be used to monitor the development of autonomous agent systems in the open source ecosystem and provide a framework to assess proprietary applications given code access.  

Although final results will require additional scale, this preliminary exercise identifies the relevance of defaults in shaping responsible behavior in the AI development value chain. The AutoGen framework constructed default set-ups for human oversight that were readily used by downstream developers. Most default set-ups were used consistently. There are exceptions - the Composio repository consistently overrode defaults to use agent code execution outside a protected docker environment. Scaling assessment can inform governance proposals to set responsible and effective defaults.

Future work can refine the code-inspection approach and subsequently scale it for all open source AutoGen applications using AI-assisted grading methods \citep{gpts}. Autonomy levels can be developed further to reflect internal dependencies among the impact and oversight attributes and to better reflect literature from other fields that uses holistic and more numerous levels of autonomy. Future work could rate autonomy levels on additional categories, such as goal setting, interactions between agent systems and fallback specifications (see Figure \ref{fig:example}), though multiple measures complementing code inspection may be needed for such attributes arising at the model layer or at inference time. Code inspection results for select systems can be compared to inference-time assessments using evaluation harnesses \citep{ukaisi}. Subsequent work can differentiate between agent applications and platforms, and generalize this approach to assess additional frameworks. 

\subsubsection*{Acknowledgments}
The views expressed in this paper are those of the authors alone and do
not necessarily reflect the official policy or position of their employers. Thanks to participants and organizers of the NeurIPS SoLaR 2024 workshop, Alan Chan, Shrey Jain, Owen Larter, and three anonymous reviewers for feedback on earlier drafts. Any errors are the authors' alone. 

\bibliography{iclr2025_conference}
\bibliographystyle{iclr2025_conference}

\newpage
\appendix
\section{Appendix}

\subsection{Levels of autonomy}
Table \ref{lit} provides a comparative overview of autonomy levels from AI and robotics literatures.

\label{autonomy-levels}
\begin{table}[hbt!]
\caption{Taxonomy of Autonomy Levels in Various Domains}
\label{lit}
\centering
\small
\begin{tabularx}{\textwidth}{@{}l*{5}{>{\raggedright\arraybackslash}X}@{}}
\toprule
\textbf{Level} & \textbf{Autonomous vehicles} \citep{ISO_SAE_22736} & \textbf{Medical robots}\citep{yang2017medical} & \textbf{AI: Human oversight} \citep{simmler2021taxonomy} & \textbf{AI: Action space} \citep{LangChainAgent2023} & \textbf{AI: Task evaluations} \citep{autonomy_evals_guide}\\
\midrule
0 & No automation & No autonomy & No autonomy & Code & -- \\
1 & Assistance: Steering-support & Assistance & Human decision: Decision-support AI & LLM call &  Few minutes \\
2 & Partial automation: Acceleration- and steering-support & Task autonomy: Human maintains discrete control & Human in the loop: Human approves action & Chain: AI-generated outputs at multiple steps, human specifying action and available actions & Several minutes (Implement simple programs) \\
3 & Conditional automation: Autonomous driving, human fallback & Conditional autonomy: Human selects plan, oversees execution & Human on the loop: Human does not veto action & Router: AI-generated steps and (single) action decision with human-determined available actions & Under an hour or few hours (Debugging etc.) \\
4 & High automation: Fallback by system & High autonomy: System makes decisions under human oversight & Human off the loop: AI takes action and then informs human & State machine: AI-generated steps and action decisions with human-determined available actions & Day long (Replicate ML papers) \\
5 & Full automation: Unlimited environment & Full automation & Human out of the loop: AI takes action independently & Autonomous: State machine with AI-generated steps and action decisions with unbounded actions & Week long or month long (Identify vulnerabilities in network, exploit them) \\
\bottomrule
\end{tabularx}
\end{table}

\subsection{Application code inspection}
Table \ref{codeflags} shows the code flags used to conduct the code inspections. Full scoring for the ten repositories can be found at: \url{https://docs.google.com/spreadsheets/d/1f7Ft24a54QapZLdAce6yIVvoyYFixarJijE39MxzMGU}.

\label{code}
\newcommand{\code}[1]{\texttt{#1}}

\begin{table}[hbt!]
\caption{Code flags for scoring autonomy of selected AutoGen applications}
\label{codeflags}
\centering
\small
\begin{tabularx}{\textwidth}{@{}l*{5}{>{\raggedright\arraybackslash}X}@{}}
\toprule
\multirow{3}{*}{\textbf{Autonomy}} & \multicolumn{2}{c}{\textbf{Impact}} & \multicolumn{3}{c}{\textbf{Oversight}} \\
\cmidrule(lr){2-3} \cmidrule(lr){4-6}
& \multicolumn{1}{c}{\makecell{Actions}} & 
\multicolumn{1}{c}{\makecell{Environment}} & 
\multicolumn{1}{c}{\makecell{Orchestration}} & \multicolumn{1}{c}{\makecell{Human-in-\\the-loop}} & \multicolumn{1}{c}{\makecell{Observability}} \\
\midrule
Main code flag & \texttt{code\_exec} \newline \texttt{ution\_config} & \texttt{use\_docker} \newline \& \newline \texttt{browser} \newline \texttt{\_config} & \texttt{max\_rounds}\newline(GroupChat) \newline or \texttt{max\_consecuti} \newline \texttt{ve\_auto\_reply} & \texttt{human\_} \newline \texttt{input\_mode} & search for \texttt{log}, \texttt{display}, and \texttt{reply\_func}\\
Lower & \texttt{=False} & \texttt{=True} \newline \& \newline not set & \texttt{<=1} or never called & \texttt{=ALWAYS} & \texttt{display\_}... \newline is configured \\
Middle & \texttt{=False} \newline \& system message with “execute the function” & \texttt{=True} \newline \& \newline set & \texttt{>1} & \texttt{=TERMINATE} \newline or mixed & Caching, logging, or \newline tracing is invoked \\
Higher & Absence of \texttt{=False} & \texttt{=False} \newline & Not set & \texttt{=NEVER} & No logging or similar \\
\bottomrule
\end{tabularx}
\end{table}

\end{document}